\documentclass[journal]{IEEEtran}

\usepackage{amsmath,amssymb}
\usepackage{graphicx}
\usepackage{booktabs}
\usepackage{cite}
\usepackage{balance}
\usepackage{microtype}
\usepackage{algorithm}
\usepackage{algpseudocode}
\usepackage[hidelinks]{hyperref}
\begin{document}

\title{Selective Tool Use for Agentic Change Visual Question Answering in Remote Sensing}

\author{
Yakoub Bazi,
Mohamed M.~Al Rahhal,
Mohamed A.~Mekhtiche,
and Mansour Zuair
\thanks{This work was supported by the Ongoing Research Funding Program, King Saud University, Riyadh, Saudi Arabia, under Grant ORF-2026-1606. Corresponding author: Yakoub Bazi.}
\thanks{Yakoub Bazi, Mohamed A.~Mekhtiche, and Mansour Zuair are with the Computer Engineering Department, College of Computer and Information Sciences, King Saud University, Riyadh 11543, Saudi Arabia (e-mail: ybazi@ksu.edu.sa; mmekhtiche@ksu.edu.sa; zuair@ksu.edu.sa).}
\thanks{Mohamed M.~Al Rahhal is with the Applied Computer Science Department, College of Applied Computer Science, King Saud University, Riyadh 11543, Saudi Arabia (e-mail: mmalrahhal@ksu.edu.sa).}
}

\maketitle

\begin{abstract}
Change visual question answering (Change VQA) requires understanding
semantic changes across bi-temporal remote sensing images. Although
vision--language models (VLMs) have shown promising performance on this
task, they remain unreliable when answering questions that require
explicit transition statistics, area measurements, or spatial
information. To address this limitation, we propose a selective
tool-use framework in which a single VLM either answers directly or
invokes a deterministic change-analysis tool to obtain question-specific
evidence. Specifically, the selected tool operates on bi-temporal
semantic maps and returns a structured observation, which the same VLM
uses to generate its final answer. To support this framework, we
construct a tool-augmented extension of CDVQA covering eight question
families and three tools for transition, spatial, and temporal analysis.
Tool-use supervision and observations are derived automatically from
the original semantic annotations, without additional manual labeling.
We then adapt Qwen3.5-4B using Low Rank Adaptation (LoRA) to jointly learn direct answering,
tool invocation, and evidence-conditioned answering. Experiments on
7,164 test questions show that selective tool use with reference
semantic maps improves overall accuracy from 73.77\% to 88.79\% and
average family accuracy from 69.11\% to 89.65\%. When the semantic maps
are predicted automatically, the framework achieves 77.47\% overall
accuracy and 75.06\% average family accuracy. These results demonstrate
the benefit of question-specific semantic evidence for Change VQA,
while highlighting the influence of semantic prediction quality on the
resulting performance. Code and tool-augmented annotations will be made
publicly available at
\url{https://github.com/yakoubbazi/ToolChangeVQA}.
\end{abstract}


\section{Introduction}

Change visual question answering (Change VQA) enables natural-language
interaction with co-registered bi-temporal remote sensing (RS) imagery.
It extends change analysis beyond identifying changed regions to
answering questions about the affected land-cover classes, their
transitions, and the extent and location of change.
CDVQA introduced a benchmark for this task, supporting
language-conditioned reasoning over bi-temporal
observations~\cite{yuan2022cdvqa}. Related studies have explored
grounded and interactive change
analysis~\cite{li2024show,deng2024changechat}, while change-captioning
methods generate natural-language descriptions of differences between
images~\cite{ricci2026changecaption}.

Recent work has shown that general-purpose vision--language models
(VLMs), when adapted to Change VQA, can achieve strong performance and
outperform several specialized
approaches~\cite{bazi2026revisiting}. However, their accuracy remains
uneven across question types. Questions about change existence or
increase and decrease can often be answered from visually apparent
differences, whereas questions involving precise proportions,
class transitions, or spatial relationships remain challenging.
For example, determining the proportion of a class that changed
requires comparing affected and total areas, while identifying its
main destination class involves comparing transitions to different
classes. Direct visual answering does not explicitly compute these
measurements, which can limit its reliability on such questions.

External analysis tools provide a means of obtaining this information
explicitly. Agentic RS systems have already combined language models
with perception and analysis
tools~\cite{liu2024changeagent,xu2024rsagent}. These developments
motivate a question for Change VQA: can a VLM learn to request
question-specific semantic evidence and use it to improve its answers?
Addressing this question requires a framework that supports both
direct answering and tool-assisted analysis, together with supervision
for selecting and using the appropriate tool.

In this paper, we propose a selective tool-use framework for
Change VQA. Given a bi-temporal image pair and a question, a single
VLM either answers directly or generates a structured call to one
deterministic change-analysis tool. The selected tool operates on
bi-temporal semantic maps and returns a structured observation,
which the same VLM uses to generate its final answer. The semantic
maps remain external to the VLM and are not provided as visual
inputs. This interaction requires at most one tool call per question
and introduces no separate routing network.

To train and evaluate the framework, we construct a tool-augmented
extension of CDVQA with eight question families and three tools for
transition, spatial, and temporal analysis. We derive supervision
from the original semantic annotations without additional manual
labeling and adapt Qwen3.5-4B using Low Rank Adaptation (LoRA)~\cite{hu2022lora} on direct-answer and
tool-assisted trajectories. Evaluation with reference and
automatically predicted semantic maps examines both the benefit
of explicit evidence and the effect of semantic prediction errors.

Our main contributions are summarized as follows:

\begin{itemize}

\item We propose a selective tool-use framework for Change VQA
in which a single VLM learns direct answering, structured tool
invocation, and evidence-conditioned answering through a unified
interaction protocol.

\item We construct a tool-augmented extension of CDVQA comprising
eight question families and three deterministic analysis tools.
Target answers, tool arguments, and
observations are derived automatically from the original
bi-temporal semantic annotations.

\item We evaluate the framework using both reference and predicted
semantic evidence. The results quantify improvements over direct
answering across question families and show how semantic prediction
quality affects the resulting Change VQA accuracy.

\end{itemize}

\section{Selective Tool-Augmented Change VQA}
\label{sec:method}
\begin{figure*}[t]
\centering
\includegraphics[width=\textwidth]{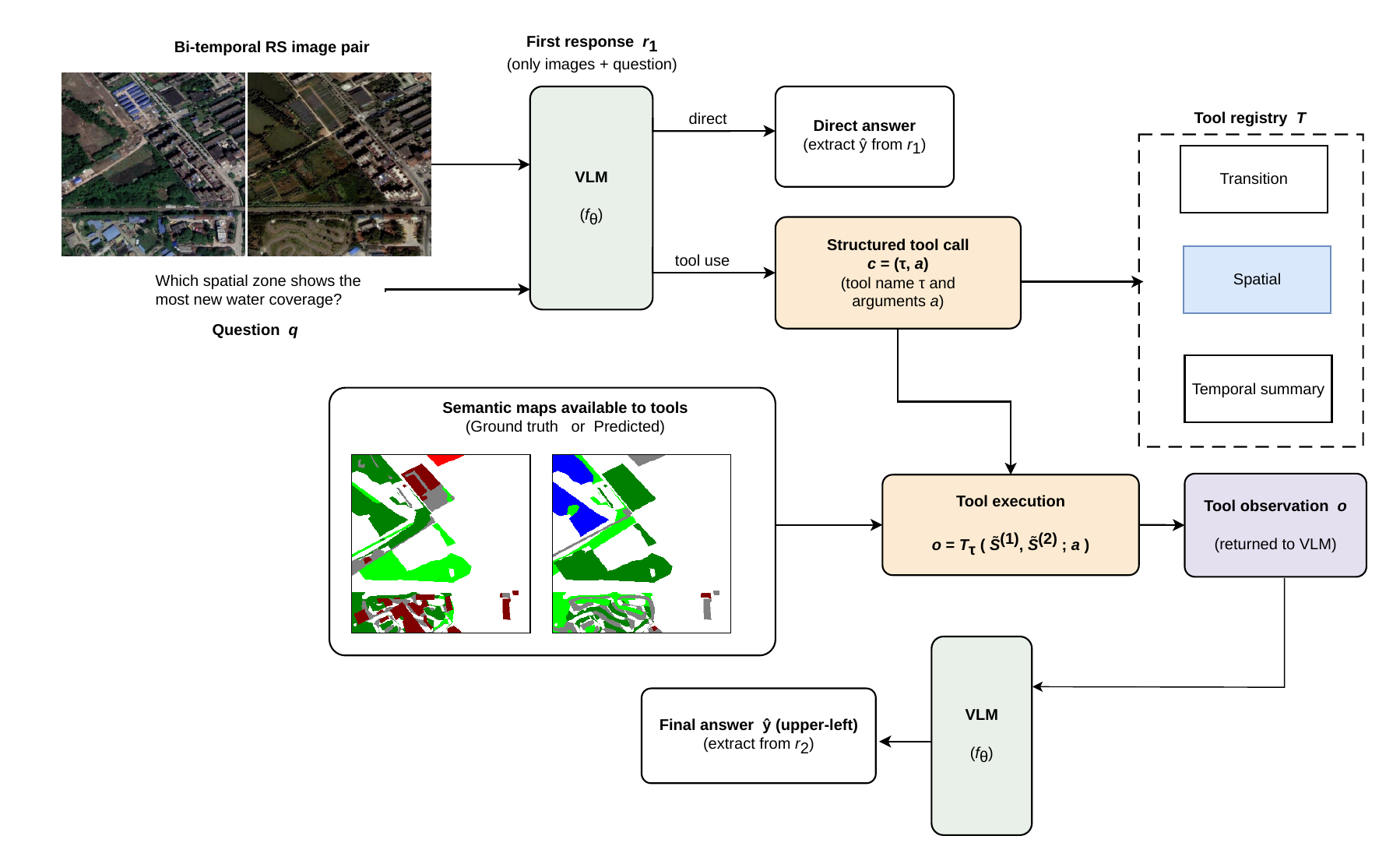}
\caption{Overview of the proposed framework. A single VLM either
answers a question directly or invokes a deterministic analysis tool.
The tool computes question-specific evidence from bi-temporal semantic
maps and returns a structured observation for final-answer generation.
The semantic maps are not provided to the VLM as visual inputs.}
\label{fig:overview}
\end{figure*}

We formulate Change VQA as a selective tool-use problem in which
a single VLM either answers directly or requests question-specific
evidence from a deterministic analysis tool. Given a bi-temporal
image pair and a question, the VLM generates either an answer or
a structured tool call. The selected tool operates on external
semantic maps and returns a structured observation, which the
same VLM uses to generate its final answer.
Fig.~\ref{fig:overview} illustrates the framework, which permits
at most one tool call per question.

\subsection{Selective Tool Use}

Let $\mathbf{I}^{(1)}$ and $\mathbf{I}^{(2)}$ denote two
co-registered RS images, $\mathbf{q}$ the input question, and
$\mathcal{T}$ the available tool registry. The VLM generates its
first response as
\begin{equation}
r_1 =
f_{\theta}
\left(
\mathbf{I}^{(1)},
\mathbf{I}^{(2)},
\mathbf{q}
\right),
\label{eq:first_response}
\end{equation}
where $f_{\theta}$ denotes the VLM with parameters $\theta$.
The interaction protocol supports two response types: a direct
answer $\hat{y}$ or a structured tool call
$c=(\tau,\mathbf{a})$, where $\tau\in\mathcal{T}$ identifies
the tool and $\mathbf{a}$ specifies its arguments.
A direct answer terminates the interaction. For a valid tool call,
the selected tool computes
\begin{equation}
\mathbf{o}
=
T_{\tau}
\left(
\widetilde{\mathbf{S}}^{(1)},
\widetilde{\mathbf{S}}^{(2)};
\mathbf{a}
\right),
\label{eq:tool_observation}
\end{equation}
where $\widetilde{\mathbf{S}}^{(1)}$ and
$\widetilde{\mathbf{S}}^{(2)}$ are the available semantic maps
at the two acquisition times, and $\mathbf{o}$ is the returned
structured observation. Only the observation is supplied to the
VLM; the semantic maps remain external.

The same VLM then generates its final answer using the original
inputs and the tool interaction:
\begin{equation}
\hat{y}
=
f_{\theta}
\left(
\mathbf{I}^{(1)},
\mathbf{I}^{(2)},
\mathbf{q},
c,
\mathbf{o}
\right).
\label{eq:final_answer}
\end{equation}
Thus, the VLM generates the response mode and tool arguments,
while the selected tool performs the requested semantic
measurements. No separate routing network is required.

\subsection{Semantic Evidence}

The tool registry contains three deterministic tools for transition,
spatial, and temporal analysis. These tools compute class-transition
and area statistics, spatial measurements of newly appearing
regions, and scene-level temporal summaries, respectively.
Their assignment to question families is described in
Section~\ref{sec:dataset}. All tools operate on bi-temporal
semantic maps in a common land-cover label space.
We consider two sources of semantic evidence. In the
\emph{reference setting}, the tools use ground-truth semantic maps.
This provides an oracle evidence source for evaluating the interaction
protocol without semantic prediction errors. It does not represent
a fully automatic image-to-answer setting.

In the \emph{predicted setting}, we train the semantic change
detection model proposed in~\cite{shen2026foundation} to
estimate bi-temporal semantic maps from the input image pair.
This model extracts multi-scale bi-temporal features and
progressively fuses them through a cascaded gated decoder
to predict the semantic state at each acquisition time.
The resulting maps are supplied to the same analysis tools,
without modifying their definitions or the VLM interaction
protocol. Accordingly, $\widetilde{\mathbf{S}}^{(1)}$ and
$\widetilde{\mathbf{S}}^{(2)}$ in
Eq.~\eqref{eq:tool_observation} denote either reference or
predicted semantic maps, depending on the evaluation setting.
The tool definitions and VLM interaction protocol remain unchanged
between these settings, allowing us to measure the effect of
replacing reference evidence with automatic predictions.
The proposed framework is not specific to the segmentation model adopted in this work as other semantic prediction models can
serve as the evidence source if they supply compatible bi-temporal
semantic maps in the required label space.

\subsection{Tool-Augmented Instruction Tuning}

We train the VLM to follow the interaction protocol
described in Eqs.~\eqref{eq:first_response}--\eqref{eq:final_answer}.
Each training example contains either a direct-answer
trajectory or a tool-assisted trajectory. The target
response mode and tool assignment follow the question-family
policy defined during benchmark construction.
For a direct-answer trajectory, the model receives the
bi-temporal images and question and is supervised to
generate the target answer. For a tool-assisted trajectory,
the first assistant response is supervised to generate
the call $c=(\tau,\mathbf{a})$. The corresponding tool
observation $\mathbf{o}$ is computed from the reference
semantic maps using Eq.~\eqref{eq:tool_observation} and
inserted into the conversation. The model is then
supervised to generate the target answer conditioned
on the original inputs, the tool call, and the observation.

Let $\mathbf{x}=(\mathbf{I}^{(1)},\mathbf{I}^{(2)},\mathbf{q})$
denote the input, and let $\mathbf{z}=(z_1,\ldots,z_L)$
denote the subsequent conversation tokens, including
assistant responses and any tool observation.
For each trajectory, we minimize the autoregressive
cross-entropy loss over assistant response tokens:
\begin{equation}
\mathcal{L}(\theta)
=
-\frac{1}{|\mathcal{A}|}
\sum_{t\in\mathcal{A}}
\log p_{\theta}
\left(z_t \mid \mathbf{x},\mathbf{z}_{<t}\right),
\label{eq:training_loss}
\end{equation}
where $\mathcal{A}$ is the set of supervised assistant
token positions, $\mathbf{z}_{<t}$ denotes the preceding
conversation tokens, and $p_{\theta}$ is the VLM's
next-token distribution. For direct-answer trajectories,
$\mathcal{A}$ contains the answer tokens. For tool-assisted
trajectories, it contains both the tool-call and final-answer
tokens. Tool-observation tokens are excluded from the loss
but remain available as context for final-answer generation.

We optimize this objective using LoRA~\cite{hu2022lora},
with trainable low-rank updates applied only to selected
attention projections in the LLM decoder. All pretrained
weights, including those of the visual encoder and
multimodal alignment modules, remain frozen.
This jointly supervises response-mode selection, tool
invocation, and answering from returned evidence.

\section{Experimental Results}
\label{sec:experiments}
\subsection{Tool-Augmented CDVQA Construction}
\label{sec:dataset}

\begin{figure}[t]
    \centering
    \includegraphics[width=0.95\columnwidth]{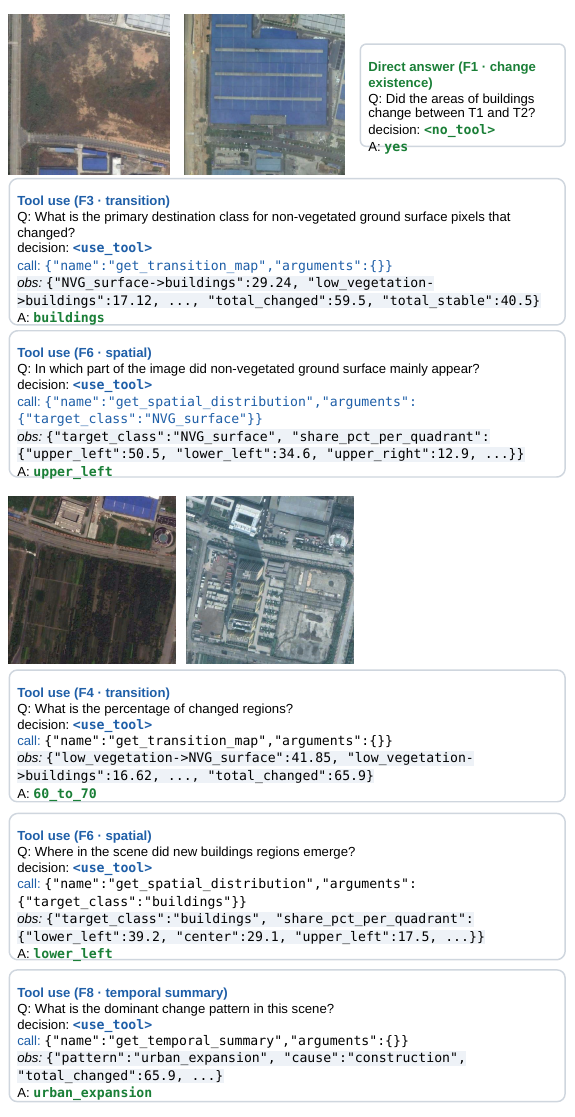}
    \caption{Representative examples from the tool-augmented CDVQA benchmark.}
    \label{fig:dataset_samples}
\end{figure}

We construct a tool-augmented extension of
CDVQA~\cite{yuan2022cdvqa} following the selective interaction protocol
defined in Section~\ref{sec:method}. Each question is assigned either
to the direct-answer path or to a tool-assisted path requiring one
deterministic tool, its arguments, and the corresponding structured
observation.

The construction uses the original co-registered bi-temporal images and
their six-class semantic annotations: non-vegetated ground surface,
buildings, playgrounds, water, low vegetation, and trees. From each
annotated image pair, we derive the target answers and the semantic
measurements required by the tools, including class transitions,
changed-area statistics, spatial occupancy, and scene-level temporal
summaries. Tool arguments and observations are generated
deterministically from these annotations, requiring no additional manual
labeling. The semantic maps are used only for benchmark construction
and tool-side evidence computation and are never provided to the VLM as
visual inputs. Representative examples are shown in
Fig.~\ref{fig:dataset_samples}.

The benchmark comprises eight question families and three deterministic
tools, as summarized in Table~\ref{tab:families}. F1--F2 follow the
direct-answer path, whereas F3--F8 require transition, spatial, or
temporal evidence. The family-to-tool mapping is fixed during dataset
construction and defines supervision for response-mode selection, tool
selection, and tool arguments.
The benchmark contains 2{,}968 co-registered image pairs and 47{,}586
questions, corresponding to an average of 16.0 questions per pair. We
partition the data at the image-pair level using a
70\%/15\%/15\% training/validation/test split, with no image pair shared
across partitions. As summarized in Table~\ref{tab:dataset_stats},
44.0\% of all questions follow the direct-answer path (F1--F2), whereas
56.0\% are tool-dependent (F3--F8), with comparable proportions across
the three splits.

\begin{table}[t]
\centering
\small
\caption{Question families and tool-use supervision policy.}
\label{tab:families}
\setlength{\tabcolsep}{3.0pt}
\begin{tabular}{cll}
\toprule
\textbf{Family} & \textbf{Question objective} & \textbf{Tool} \\
\midrule
F1 & change existence & direct \\
F2 & increase/decrease & direct \\
F3 & destination class & transition \\
F4 & changed/stable proportion & transition \\
F5 & largest/smallest change & transition \\
F6 & location of new regions & spatial \\
F7 & pairwise change comparison & transition \\
F8 & dominant temporal pattern & temporal summary \\
\bottomrule
\end{tabular}
\end{table}

\begin{table}[t]
\centering
\small
\caption{Statistics of the tool-augmented CDVQA benchmark. Direct and
Tool denote direct-answer (F1--F2) and tool-dependent (F3--F8)
questions, respectively.}
\label{tab:dataset_stats}
\setlength{\tabcolsep}{4.0pt}
\begin{tabular}{lccccc}
\toprule
\textbf{Split} & \textbf{Pairs} & \textbf{Questions}
& \textbf{Direct} & \textbf{Tool} & \textbf{Q/pair} \\
\midrule
Train & 2{,}077 & 33{,}299 & 14{,}621 & 18{,}678 & 16.0 \\
Val   & 445     & 7{,}123  & 3{,}132  & 3{,}991  & 16.0 \\
Test  & 446     & 7{,}164  & 3{,}194  & 3{,}970  & 16.1 \\
\midrule
\textbf{Total} & \textbf{2{,}968} & \textbf{47{,}586}
& \textbf{20{,}947} & \textbf{26{,}639} & \textbf{16.0} \\
\bottomrule
\end{tabular}
\end{table}

\subsection{Experimental Setup}
\label{sec:exp_setup}

We evaluate final answers using exact-match accuracy and report
per-family accuracy, overall accuracy (OA), and average family accuracy
(AA):

\begin{equation}
\mathrm{OA}
=
\frac{1}{N}
\sum_{i=1}^{N}
\mathbf{1}\!\left[\hat{y}_i=y_i\right],
\qquad
\mathrm{AA}
=
\frac{1}{8}
\sum_{k=1}^{8}
\mathrm{Acc}_{F_k},
\label{eq:metrics}
\end{equation}

where $N$ is the number of test questions and
$\mathrm{Acc}_{F_k}$ denotes the accuracy for question family $F_k$.

For the VLM $f_{\theta}$ We use Qwen3.5-4B~\cite{qwen2026native}, a native
vision--language model with a 4-billion-parameter. The model is pretrained through early fusion of
visual and textual tokens and uses a hybrid decoder combining
Gated DeltaNet and full-attention layers to process multimodal
inputs jointly. We adapt this model using
LoRA~\cite{hu2022lora}, with adapters applied only to selected
attention projections within the LLM decoder. The visual
encoder, multimodal alignment modules, and all pretrained
weights remain frozen; only the LoRA parameters are updated.
We set the LoRA rank to $r=16$ and the scaling parameter to
$\alpha=32$. Training is performed for three epochs using
AdamW with a learning rate of $5\times10^{-5}$, cosine
learning-rate decay, a warmup ratio of 0.1, and weight decay
of 0.01. The experiments are conducted on a single
NVIDIA RTX A6000 GPU.

\subsection{Change VQA Results}
\label{sec:main_results}

Table~\ref{tab:main_results} compares direct answering with
selective tool use under reference and predicted semantic
evidence. The reference setting assesses performance with
ground-truth semantic maps, whereas the predicted setting
evaluates the complete pipeline using maps estimated from
the input images.

\begin{table*}[t]
\centering
\footnotesize
\caption{Change VQA accuracy (\%) on the test questions.
Ref. and Pred. denote reference and predicted semantic maps,
respectively. Bold values indicate the highest accuracy
in each column.}
\label{tab:main_results}
\setlength{\tabcolsep}{2.2pt}
\renewcommand{\arraystretch}{1.08}
\begin{tabular}{llcccccccccc}
\toprule
\textbf{Method} & \textbf{Evidence}
& \textbf{F1} & \textbf{F2} & \textbf{F3} & \textbf{F4}
& \textbf{F5} & \textbf{F6} & \textbf{F7} & \textbf{F8}
& \textbf{OA} & \textbf{AA} \\
\midrule
Direct answer & -- &
\textbf{88.60} & \textbf{84.74} &
69.06 & 50.34 & 47.20 &
52.42 & 80.17 & 80.38 &
73.77 & 69.11 \\

Selective tool use & Pred. &
88.36 & 82.06 &
74.50 & 63.40 & 48.69 &
66.46 & 83.51 & 93.46 &
77.47 & 75.06 \\

Selective tool use & Ref. &
88.36 & 82.06 &
\textbf{97.77} & \textbf{99.83} & \textbf{59.51} &
\textbf{99.76} & \textbf{89.87} & \textbf{100.00} &
\textbf{88.79} & \textbf{89.65} \\
\bottomrule
\end{tabular}
\end{table*}

Direct answering achieves 73.77\% OA and 69.11\% AA,
with its highest accuracies on F1 and F2. Performance is
considerably lower on F4, F5, and F6, reaching 50.34\%,
47.20\%, and 52.42\%, respectively. These families involve
proportion estimation, comparison of change quantities,
and spatial localization, highlighting the difficulty of
obtaining precise measurements through direct visual answering.
With reference semantic evidence, selective tool use achieves
88.79\% OA and 89.65\% AA, improving over direct answering
by 15.02 and 20.54 percentage points, respectively.
Improvements occur across all six tool-assigned families.
In particular, F3 increases from 69.06\% to 97.77\%,
F4 from 50.34\% to 99.83\%, and F6 from 52.42\% to
99.76\%, while F8 reaches 100.00\%. These results show
that access to reference semantic evidence substantially
improves answers involving class transitions, change
proportions, spatial locations, and temporal patterns.

F5 remains the most challenging tool-dependent family, improving from
47.20\% to 59.51\% even with reference evidence. To separate tool
invocation from downstream answering, we further inspect the
3{,}970 tool-dependent test questions. The model follows the predefined
tool-use policy and executes the required tool correctly, while the
evidence-conditioned final-answer accuracy reaches 91.56\% overall.
This indicates that most residual errors occur after evidence
acquisition rather than during tool selection or execution. F5 is
particularly sensitive to this stage because answering requires
comparing and ranking several transition quantities rather than
interpreting a single measurement.
\subsection{Predicted Semantic Evidence}
\label{sec:predicted_evidence}

We next examine whether selective tool use improves over
direct answering when semantic evidence is predicted
automatically. We use the semantic change detection
model in~\cite{shen2026foundation} to generate the
bi-temporal maps. Table~\ref{tab:segmenter_results}
reports the predictor's performance using the evaluation
protocol in~\cite{shen2026foundation}: pixel-level overall
accuracy (OA), mean intersection over union (mIoU),
separated kappa (SeK), and semantic change detection
F1 score ($F_{\mathrm{scd}}$).
Using the predicted maps, selective tool use achieves
77.47\% OA and 75.06\% AA on Change VQA, compared with
73.77\% OA and 69.11\% AA for direct answering
(Table~\ref{tab:main_results}). These improvements of
3.70 and 5.95 percentage points show that the benefit
of semantic evidence persists in the complete automatic
pipeline, despite errors in the predicted maps.

\begin{table}[t]
\centering
\small
\caption{Semantic change detection performance on the test
set (\%). mIoU: mean intersection over union;
SeK: separated kappa; $F_{\mathrm{scd}}$: semantic change
detection F1 score; OA: pixel-level overall accuracy.}
\label{tab:segmenter_results}
\setlength{\tabcolsep}{4pt}
\begin{tabular}{lcccc}
\toprule
\textbf{Model} & \textbf{mIoU} & \textbf{SeK} &
\textbf{$F_{\mathrm{scd}}$} & \textbf{OA} \\
\midrule
CD Segmentation model~\cite{shen2026foundation} &
73.08 & 25.42 & 65.29 & 87.38 \\
\bottomrule
\end{tabular}
\end{table}

The improvements extend across all six tool-assigned
families. The largest gains occur in F6, which increases
from 52.42\% to 66.46\%, F8, from 80.38\% to 93.46\%,
and F4, from 50.34\% to 63.40\%. Thus, predicted evidence
is particularly useful for spatial localization,
temporal-pattern identification, and proportion estimation.
F3 and F7 also improve, from 69.06\% to 74.50\% and
from 80.17\% to 83.51\%, respectively. The improvement
in F5 is smaller, from 47.20\% to 48.69\%, indicating
that identifying the largest or smallest change remains
challenging. F1 and F2 remain below
the direct-answer baseline; the aggregate improvement
therefore comes from the tool-assigned families.

Comparison with reference evidence reveals the remaining
sensitivity to semantic prediction quality. OA decreases
from 88.79\% with reference maps to 77.47\% with predicted
maps, a gap of 11.32 percentage points. F4 and F6 show
particularly large reductions, from 99.83\% to 63.40\%
and from 99.76\% to 66.46\%, respectively. Nevertheless,
both remain substantially above their direct-answer
baselines. F8 retains 93.46\% accuracy compared with
100.00\% under reference evidence, showing a smaller
reduction on this benchmark. 
These results suggest that semantic maps predicted by the
method in~\cite{shen2026foundation} provide useful evidence
for Change VQA. The observed gap from reference evidence
also suggests that the gains depend on the quality of
the semantic maps produced by the chosen prediction method.

\section{Conclusion}

We proposed a selective tool-use framework for Change VQA
in which a single VLM either answers directly or invokes
a deterministic tool to obtain question-specific evidence
from bi-temporal semantic maps. We constructed a
tool-augmented CDVQA benchmark with eight question families
and three analysis tools, using supervision derived
automatically from existing annotations. Experiments with
reference and predicted semantic maps show higher overall
accuracy than direct answering, with improvements across
the tool-assigned question families. Questions about the
largest or smallest change remain challenging, even with
reference maps. Future directions include improving answers
to these comparative questions and investigating alternative
segmentation models to assess how evidence quality affects
performance. The framework could also be explored in other
RS tasks that benefit from explicit semantic
measurements.
\balance
\bibliographystyle{IEEEtran}
\bibliography{refs_grsl_agentic_v2}

\end{document}